\documentclass[10pt,twocolumn,letterpaper]{article}

\usepackage{cvpr}
\usepackage{times}
\usepackage{epsfig}
\usepackage{graphicx}
\usepackage{amsmath}
\usepackage{amssymb}
\usepackage{color}

\newcommand{\Rd}{{\mathbb R}}

\newcommand{\Xc}{{\mathcal X}}
\newcommand{\Yc}{{\mathcal Y}}
\newcommand{\Fc}{{\mathcal F}}

\usepackage[pagebackref=true,breaklinks=true,letterpaper=true,colorlinks,bookmarks=false]{hyperref}

 \cvprfinalcopy 

\def\cvprPaperID{3538} 

\ifcvprfinal\fi
\begin{document}

\title{Deep Residual Learning for Compressed Sensing CT Reconstruction \\ via Persistent Homology Analysis }

\author{Yo Seob Han,  Jaejun Yoo and Jong Chul Ye\\
Bio Imaging Signal Processing Lab\\
Korea Ad. Inst. of Science \& Technology (KAIST) \\
291 Daehak-ro, Yuseong-gu, Daejeon 34141, Korea
\\
{\tt\small \{hanyoseob,jaejun2004,jong.ye\}@kaist.ac.kr}
}

\maketitle

\begin{abstract}
Recently, compressed sensing  (CS) computed tomography (CT) using sparse projection views has been
extensively investigated to reduce the potential risk of radiation to patient.
However, due to the insufficient number of projection views, an analytic reconstruction approach results in  severe streaking artifacts
and CS-based iterative approach is computationally very expensive.
To address this issue, here  we propose a novel deep residual learning approach for sparse-view CT reconstruction. 
%
%
%
Specifically, based on a novel persistent homology analysis showing that the manifold of streaking artifacts is topologically simpler than original one,  a deep residual learning
architecture that estimates the streaking artifacts is developed. Once a streaking artifact image is estimated, an artifact-free image can be obtained by subtracting the streaking artifacts from the input image.
Using extensive experiments with  real patient data set, we confirm that the proposed residual learning provides significantly better image reconstruction performance with several orders of magnitude
faster computational speed. 

%
\end{abstract}

\section{Introduction}


Recently,  deep learning approaches have achieved tremendous
success in  classification problems~\cite{krizhevsky2012imagenet} as well as low-level computer vision problems such as segmentation~\cite{ronneberger2015u}, denoising~\cite{zhang2016beyond}, super resolution~\cite{kim2015accurate, shi2016real}, etc. 
The theoretical origin of their success  has been investigated by a few authors \cite{poole2016exponential,telgarsky2016benefits}, where
the exponential expressivity under a given network complexity (in terms of VC dimension \cite{anthony2009neural} or Rademacher complexity \cite{bartlett2002rademacher})
has been attributed to their success.

In medical imaging area,  there have been also extensive research activities applying deep learning.   However, most of these works are
focused on  image-based diagnostics, and 
their applications to image reconstruction problems such as X-ray computed tomography (CT)  reconstruction  is relatively less investigated.

In X-ray CT,
due to the potential risk of radiation exposure, 
the main research thrust is  to reduce the  radiation dose.  Among various approaches for  low-dose CT, sparse view CT is a recent proposal that reduces the
radiation dose by reducing the number of projection views \cite{sidky2008image}. 
However, due to the insufficient 
projection views,  standard reconstruction using the
 filtered back-projection (FBP) algorithm exhibits severe streaking artifacts.  Accordingly,  researchers have extensively employed 
compressed sensing approaches   \cite{donoho2006compressed} that minimize the total variation (TV) or other sparsity-inducing penalties under the data fidelity  \cite{sidky2008image}. 
These approaches are, however, computationally very expensive due to the repeated applications of projection and back-projection during iterative update steps.
 
\begin{figure*}[t]
    \centerline{\includegraphics[width=0.95\linewidth]{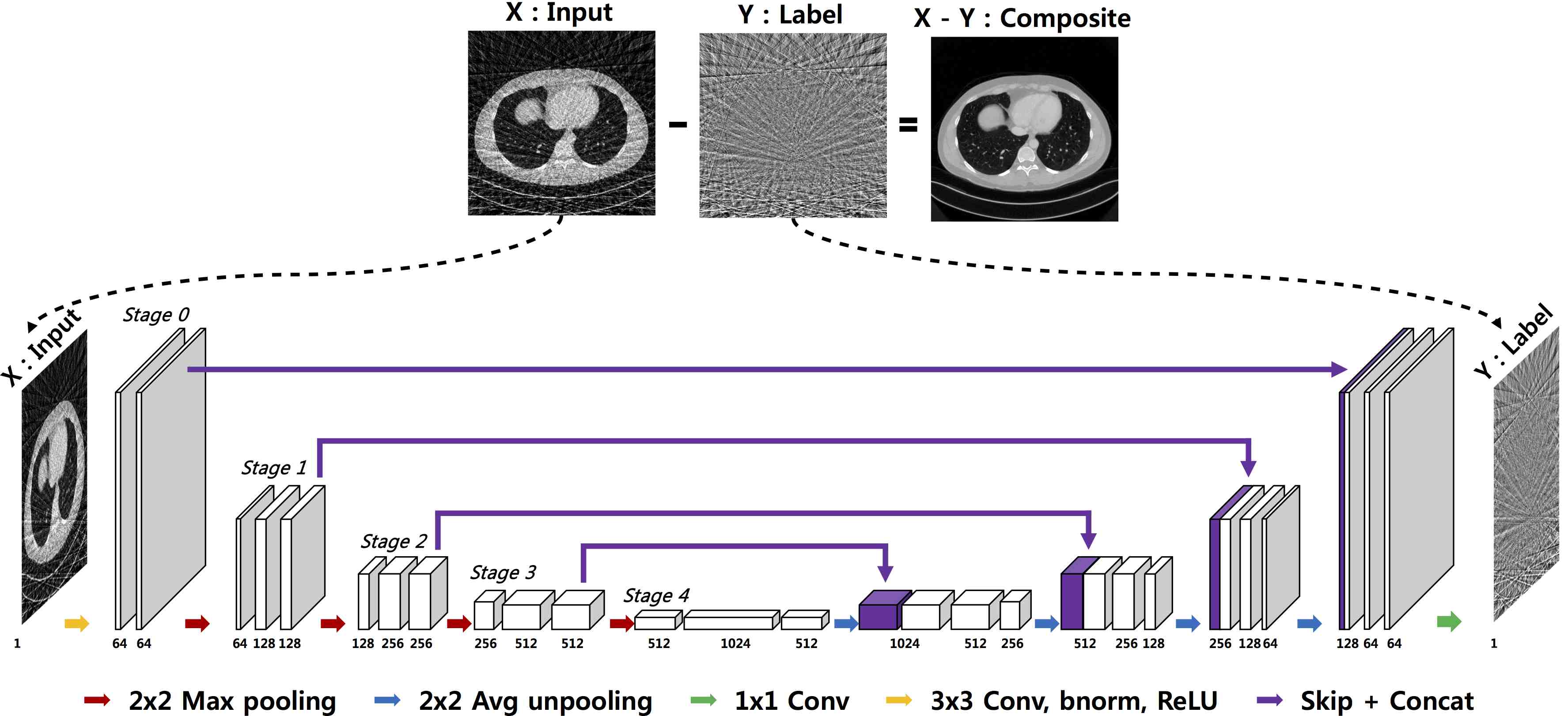}}
    \caption{The proposed deep residual learning architecture for sparse view CT reconstruction. }
    \label{fig:proposed_network}
\end{figure*}

Therefore,   the main goal of this paper is to develop a novel deep  CNN architecture for sparse view CT reconstruction that outperforms the
existing approaches in its computational speed as well as reconstruction quality. 
However, a direct application of conventional CNN architecture turns out to be inferior, because X-ray CT images  have high texture details that are often difficult to estimate from sparse view reconstructions.
To address this,  we propose a  novel {\em deep  residual learning } architecture to learn streaking artifacts.
Once the streaking artifacts are estimated, an artifact-free image is then obtained by subtracting the estimated streaking artifacts  as shown in Fig.~\ref{fig:proposed_network}.

The proposed deep residual learning  is based on our conjecture  that streaking artifacts from sparse view CT reconstruction may have simpler topological structure such that learning streaking artifacts
is easier than learning  the 
original artifact-free images. 
To prove this conjecture, we employ a recent computational topology tool called  the {\em persistent homology} \cite{edelsbrunner2008persistent} to
show that the residual manifold is much simpler than the original one.
For practical implementation, we investigate several  architectures of residual learning,  which consistently shows that residual learning is 
better than image learning.  In addition, among  various residual learning architecture,  we show
that multi-scale deconvolution network  with contracting path - which is often called 
U-net structure~\cite{ronneberger2015u}  for image segmentation - 
is most effective in removing streaking artifacts especially from very sparse number of projection views. 

\noindent {\bf Contribution: } 
 In summary, our contributions are as following.  First, a computational topology tool called the persistent homology is proposed as a novel
 tool to analyze  the manifold of the label data.  The analysis clearly shows the advantages of the residual learning in sparse view CT reconstruction.
 Second,  among various type of residual learning architecture,  multi-scale architecture known as U-net is shown the most effective.  We show that the advantage of this architecture is originated from its enlarged receptive fields 
 that can  easily capture  globally distributed artifact patterns.  Finally,  to our best knowledge,  the proposed algorithm is the first 
 deep learning architecture that successfully reconstruct high resolution images from very sparse number of projection views.
 Moreover, the proposed method significantly outperforms the existing compressed sensing CT approach in both image quality and reconstruction speed.

\section{Related works}

In CT reconstruction, only a few deep learning architectures are available.
Kang \etal\cite{kang2016deep} provided the first systematic study of deep CNN 
in low-dose CT from  reduced X-ray tube currents and showed that a deep CNN using  directional wavelets is more efficient in removing
low-dose related CT noises. 
Unlike these low-dose artifacts  originated from reduced tube current,
the streaking artifacts from sparse projection views exhibit globalized artifact patterns, which is difficult to remove
using conventional denoising CNNs \cite{chen2015learning, mao2016image, xie2012image}. 
Therefore, to our best knowledge, there exists no deep learning architecture for sparse view CT reconstruction.

The residual learning concept was first introduced by He \etal\cite{he2015deep} for image recognition.
In low-level computer vision problems,  Kim \etal\cite{kim2015accurate} employed a residual learning for  a super-resolution (SR) method. 
In these approaches, the residual learning was implemented by a skipped connection corresponding to an identity mapping.
Unlike these architectures,  Zhang \etal\cite{zhang2016beyond} proposed a direct residual learning architecture for image denoising and super-resolution, which has inspired our method.

The proposed architecture in Fig.~\ref{fig:proposed_network} is originated from U-Net developed by Ronneberger \etal\cite{ronneberger2015u} for image segmentation.
This architecture was motivated from another deconvolution network for image segmentation by Noh \etal\cite{noh2015learning} by adding contracting path and pooling/unpooling layers.
However, we are not aware of any prior work that employed this architecture beyond the image segmentation.

\section{Theoretical backgrounds}

Before we explain the proposed deep residual learning architecture,  this section provides
theoretical backgrounds. 

\subsection{Generalization bound}

In a learning problem, based on a random  observation (input)
$X \in \Xc$ and a label $Y \in \Yc$ generated by a distribution $D$, we are interested in  estimating a  regression function $f: X \rightarrow Y$ in a functional space $\Fc$
that minimizes the  risk
$L(f) = E_D \|Y-f(X)\|^2.$ 
A major technical issue 
 is that the associated probability distribution $D$ is unknown.
Moreover, we only have a finite
 sequence of independent and identically distributed training data 
$S=\{(X_1,Y_1),\cdots, (X_n, Y_n)\}$
such that only 
 an empirical risk
$\hat L_n(f) = \frac{1}{n}\sum_{i=1}^n \|Y_i - f(X_i)\|^2$
is available.
Direct minimization of empirical risk is, however,  problematic due to the  overfitting.

To address these issues, statistical learning theory  \cite{anthony2009neural} {\color{black}has} been developed to 
bound the risk of a learning algorithm in terms of complexity measures (eg. VC dimension and shatter coefficients) and the empirical
risk. Rademacher complexity \cite{bartlett2002rademacher} is one of the {\color{black}most} modern notions of complexity that is distribution dependent
and defined for any class of real-valued functions.
Specifically, with probability $\geq 1-\delta$,  for every  function $f\in \Fc$, 
\begin{equation}\label{eq:L}
L(f) \leq \underbrace{\hat L_n(f)}_{\text{empirical risk}} + \underbrace{2 \hat R_n(\Fc)}_{\text{complexity penalty}}+ 3 \sqrt{\frac{\ln(2/\delta)}{n}}
\end{equation}
where the empirical Rademacher complexity $\hat R_n(\Fc)$ is defined to be
$$\hat R_n(\Fc)=  E_\sigma \left[\sum_{f\in \Fc} \left(\frac{1}{n}\sum_{i=1}^n \sigma_i f(X_i) \right) \right],$$ 
where $\sigma_1,\cdots, \sigma_n$ are independent random variables uniformly chosen  from $\{-1,1\}$.
Therefore, to reduce the risk,  we need to minimize  both the empirical risk (i.e.  data fidelity) and the complexity term in \eqref{eq:L} simultaneously.

In neural network,  empirical risk is determined by the representation power of a network \cite{telgarsky2016benefits}, whereas the complexity term is determined by
the structure of a network. 
Furthermore, it was shown  that the capacity of  implementable functions grows exponentially with respect to the number of hidden units \cite{poole2016exponential,telgarsky2016benefits}. 
Once the network architecture is determined,  its capacity is fixed. Therefore,
the  performance of the network is now dependent on the complexity of the manifold of label $Y$ that a given deep network tries to approximate. 

In the following, using the persistent homology analysis, we show that the residual manifold composed of X-ray CT streaking artifacts is topologically simpler than the original one.

\subsection{Manifold of  CT streaking artifacts}


\begin{figure}[t]
    \centerline{\includegraphics[width=0.95\linewidth]{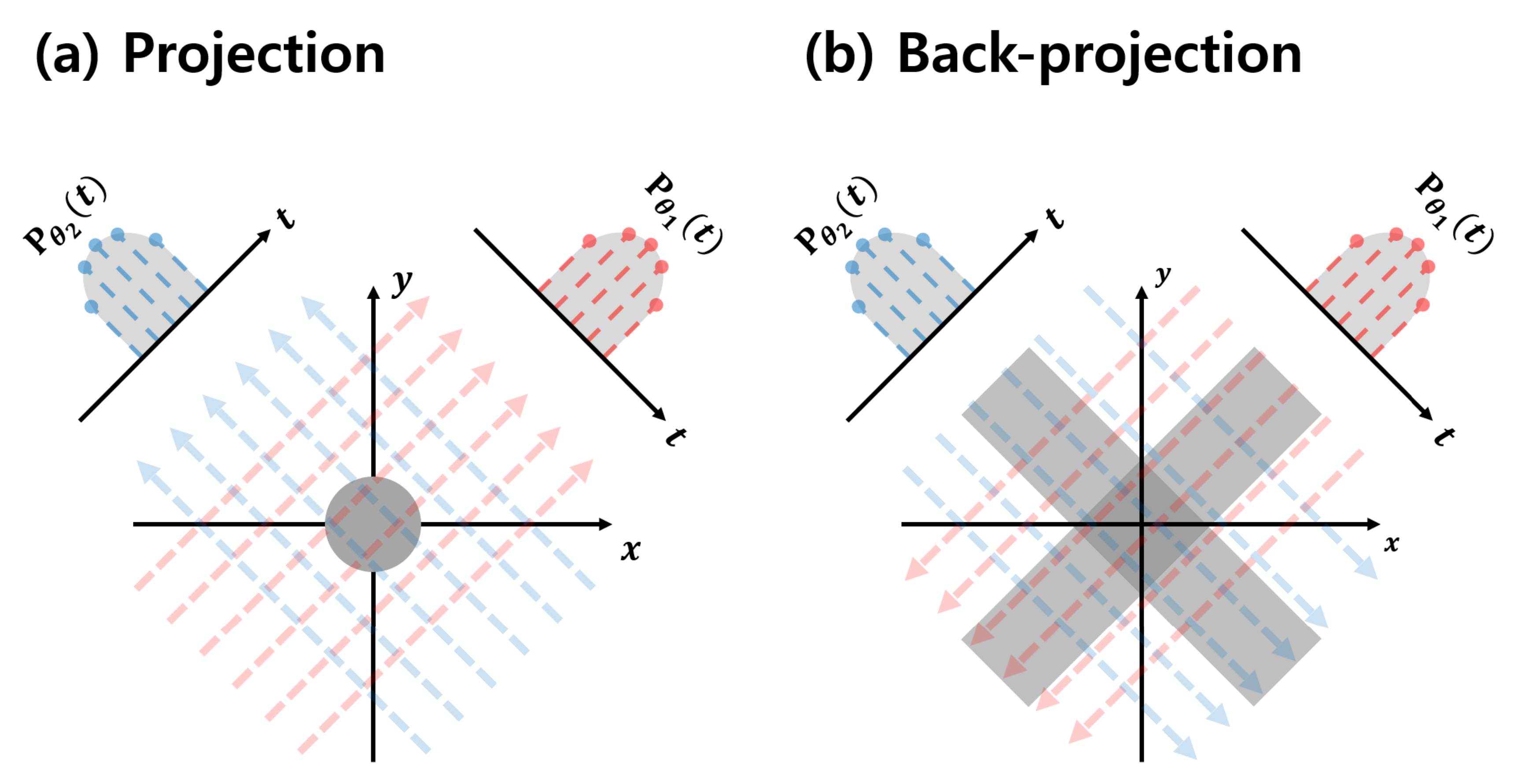}}
    \caption{CT projection and back-projection operation.}
    \label{fig:ct_system}
\end{figure}

In order to describe the manifold of CT streaking artifacts,
this section starts with a brief introduction of CT physics and its analytic reconstruction method.
For simplicity,
a parallel-beam CT system is described.  In CT, an X-ray photon undergoes attenuation according to
Beers-Lambert law while it passes through the body. Mathematically, this can be described by a Radon transform.
Specifically, the projection 
measurement at the detector distance $t$ in the projection angle $\theta$ is 
described by 
\begin{eqnarray*}\label{eq:proj}
P_\theta(t)
&=&\int_{-\infty}^{\infty}\int_{-\infty}^{\infty}{f(x,y)\delta(t - x\rm{cos}\theta - y\rm{sin}\theta)}dxdy,
\end{eqnarray*}
where $f(x,y)$ denotes the underlying image, and $t = x\rm{cos}\theta + y\rm{sin}\theta$ denotes the X-ray propagation path as shown in Fig.~\ref{fig:ct_system}(a).
If densely sampled  projection measurements are available, 
the filtered back-projection (FBP)
\begin{eqnarray}\label{eq:flt_back_proj}
f(x,y)
&=&\int_{0}^{\pi}d\theta\int_{-\infty}^{\infty}{|\omega|P_\theta(\omega)e^{j2\pi\omega t}}d\omega,
\end{eqnarray}
 becomes the inverse Radon transform,
 where $|\omega|$ denotes ramp filter and $P_\theta(\omega)$ indicates 1-D Fourier transform of projection along the detector $t$.

%
%

\begin{figure}[t]
    \centerline{\includegraphics[width=0.95\linewidth]{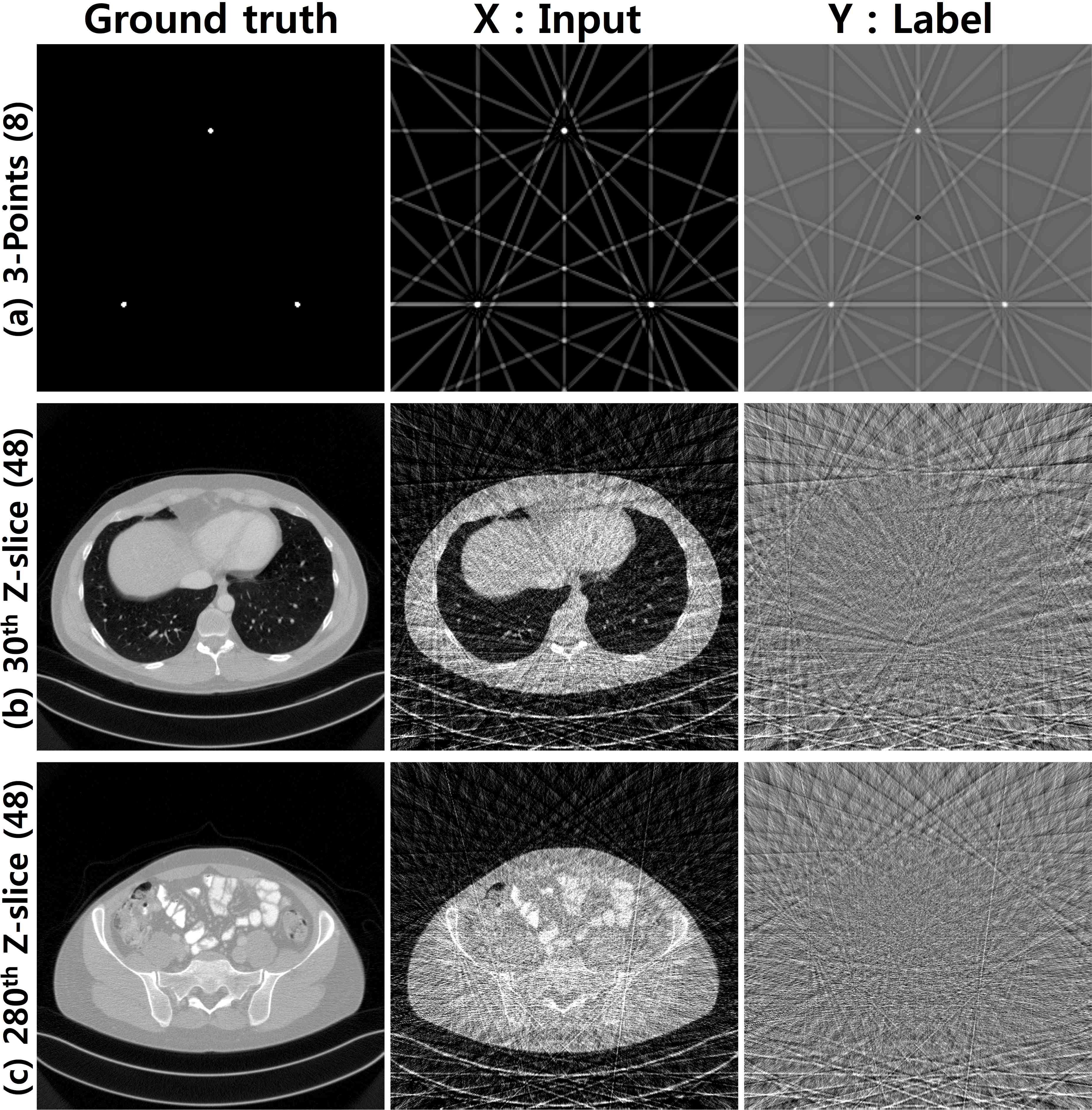}}
    \caption{CT streaking artifact patterns for  (a) three point targets from 8 view projection measurements and
    (b)(c)  reconstruction images  from 48 projections.  }
    \label{fig:streaking_pattern}
\end{figure}

 In \eqref{eq:flt_back_proj}, 
 the outer integral corresponds to the back-projection  that projects the filtered sinogram back along the original
 X-ray beam propagation direction (see Fig.~\ref{fig:ct_system}(b)).
Accordingly, if the number of projection view is not sufficient,  this introduces  streaking artifacts.
Fig. \ref{fig:streaking_pattern}(a) shows a FBP reconstruction result  from eight view projection measurements for  three point targets.
There exist significantly many streaking artifacts radiating from each point target.
Fig. \ref{fig:streaking_pattern}(b)(c) show  two reconstruction images and their artifact-only images when only 48 projection views are available.
Even though the underlying images are very different from the point targets and from each other, similar streaking artifacts radiating from objects are consistently observed.
This suggests that the streaking artifacts from different objects may have similar topological structures. 

\begin{figure}[!hbt]
    \centerline{\includegraphics[width=1.0\linewidth]{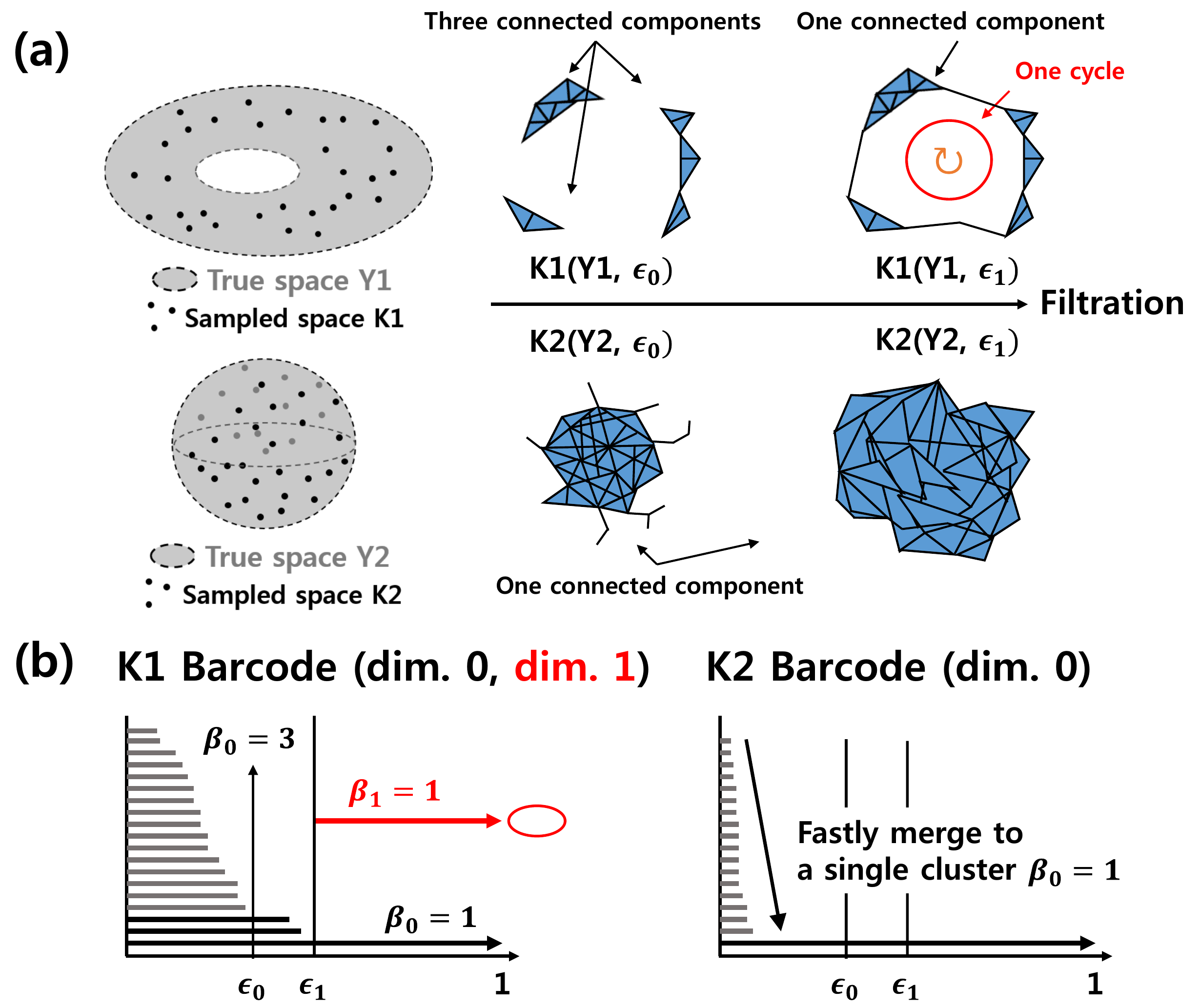}}
    \caption{(a) Point cloud data $K$ of true space $Y$ and its configuration over $\epsilon$ distance filtration. $Y_1$ is a doughnut and $Y_2$ is a sphere shaped space each of which represents a complicated space and a simpler space, respectively. (b) Zero and one dimensional barcodes of $K1$ and $K2$. Betti number can be  easily calculated by counting the number of barcodes at each filtration value $\epsilon$.  }
    \label{fig:topo}
\end{figure}

The complexity of a manifold is a topological concept. Thus,  it should be analyzed using topological tools.
In algebraic topology, Betti numbers ($\beta_m$) represent the number of $m$-dimensional holes of a manifold. For example, $\beta_0$ and $\beta_1$ are the number of connected components and cycles, respectively. They are frequently used to investigate the characteristic of underlying data manifold \cite{edelsbrunner2008persistent}. 
Specifically, we can infer the topology of a data manifold by  varying the similarity measure between the data points and tracking the changes of Betti numbers. 
As allowable distance $\epsilon$ increases, point clouds merge together and finally become a single cluster. 
Therefore, the point clouds with high diversity will merge slowly and this will be represented as a slow decrease in Betti numbers. 
For example, in Fig.~\ref{fig:topo}(a), the space $Y_1$ is a doughnut with a hole (i.e. $\beta_0=1$ and $\beta_1=1$) whereas
$Y_2$ is a sphere-like cluster (i.e. $\beta_0=1$ and $\beta_1=0$). 
Accordingly,  $Y_1$ has longer zero dimensional \emph{barcodes} persisting over $\epsilon$ in Fig.~\ref{fig:topo}(b). In other words, it has a persisting one dimensional barcode implying the distanced configuration of point clouds that cannot be overcome until they reach a large $\epsilon$.
This  persistence of Betti number is an important topological characteristics, and the
recent {\em persistent homology} analysis utilizes this to investigate the topology of  data  \cite{edelsbrunner2008persistent}.

In  Bianchini \etal. \cite{bianchini2014complexity},   Betti number of 
the set of inputs scored with a nonnegative
response was used as a capacity measure of a deep neural network.
However, our approach is novel, because we are interested in investigating the 
complexity of the label manifold. 
As will be shown in Experiment, the persistent homology analysis clearly show
that the residual manifold has much simpler topology than the original one.

\begin{figure}[!hbt]
    \centerline{\includegraphics[width=0.95\linewidth]{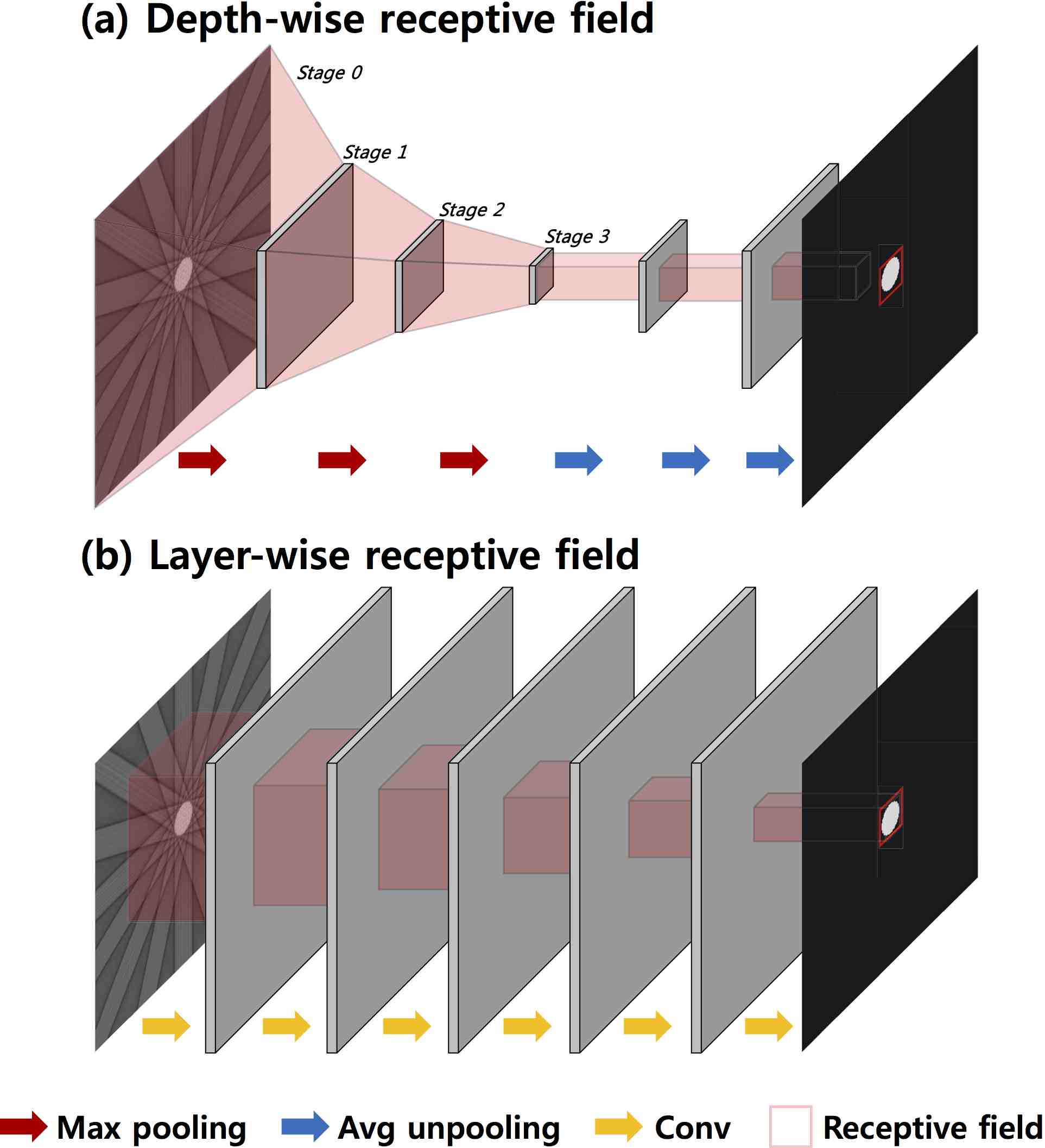}}
    \caption{Effective receptive field comparison.}
    \label{fig:artifact_pattern}
\end{figure}

%

\section{Residual Learning Architecture}



As shown in Fig. \ref{fig:proposed_network},
the proposed residual network  consists of convolution layer, batch normalization~\cite{ioffe2015batch}, rectified linear unit (ReLU) \cite{krizhevsky2012imagenet}, and contracting path connection with concatenation~\cite{ronneberger2015u}.
Specifically,
 each stage contains  four sequential layers composed of convolution with $3\times3$ kernels, batch normalization, and ReLU layers.
 Finally,  the last stage has two sequential layers and the last layer contains only one convolution layer with $1\times1$ kernel.
In the first half of the network, each stage is followed by a max pooling layer, whereas an average unpooling layer
is used in
the later half of the network.
Scale-by-scale contracting paths are used to concatenate the results from the front part of the network to the later part of network. 
The number of channels for each convolution layer is illustrated in  Fig. \ref{fig:proposed_network}.
Note that the number of channels are doubled after each pooling layers.


Fig. \ref{fig:artifact_pattern} compares  the network depth-wise effective receptive field for a simplified form of the proposed network and a reference network without pooling layer.
With the same convolutional filter, the effective receptive field is enlarged in the proposed architecture.
Considering that the streaking artifact has globally distributed pattern as illustrated in Fig.~\ref{fig:streaking_pattern},
the enlarged effective receptive field from the multi-scale residual learning is more advantageous in removal of the streaking artifacts.

\section{Experimental Results}

\subsection{Data Set}
As a training data, we used the nine patient data  provided by AAPM Low Dose CT Grand Challenge
(http://www.aapm.org/GrandChallenge/LowDoseCT/). 
The data is composed of 3-D CT projection data from  2304 views.
Artifact-free original images were generated by FBP using all  2304 projection views. 
Sparse view CT reconstruction input images    $X$ were generated using FBP from 48, 64, 96, and 192 projection views, respectively.
For the proposed residual learning,  the label data $Y$  were defined as the difference between the sparse view reconstruction and the full-view reconstruction.

Among the nine patient data,  eight patient data were used for training, whereas a test was conducted using the remaining one patient data. 
This corresponding to  3602 slices of $512\times 512$ images for the training data, and 488 slices of  $512\times 512$ images for the test data.
The training data was augmented by conducting horizontal and vertical flipping.
For the training data set,  we used the FBP reconstruction using  both 48 and 96 projection views as input $X$ and the difference between the full-view (2304views) reconstruction and the sparse view reconstructions were used as label $Y$.

\begin{figure}[!b]
    \centerline{\includegraphics[width=1.0\linewidth]{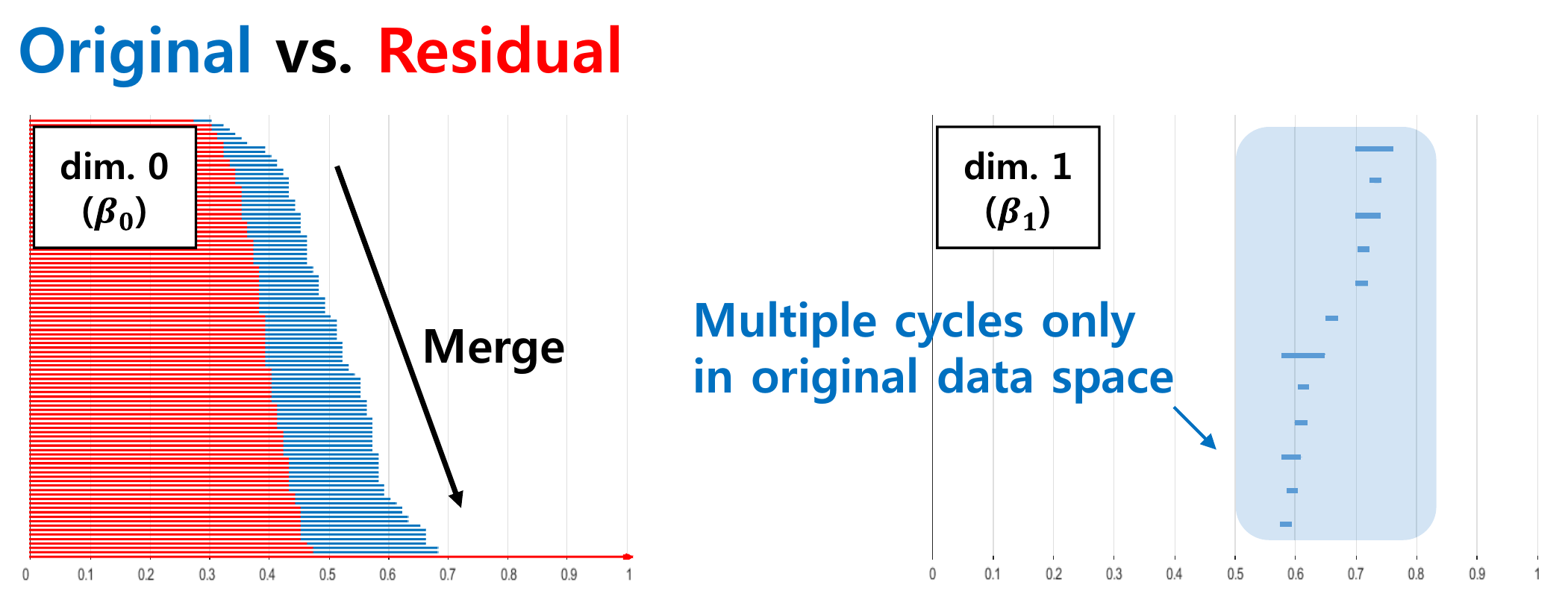}}
    \caption{Zero and one dimensional barcodes of the artifact-free original CT images (blue) and streaking artifacts (red). }
    \label{fig:topo_exp}
\end{figure}

\subsection{Persistent homology analysis}

To compare the topology of the original and residual image spaces, we calculated Betti numbers using a toolbox called JAVAPLEX (http://appliedtopology.github.io/ javaplex/). Each label image of size $512\times 512$ was set to  a point in $\Rd^{512^2}$ vector space to generate a point cloud. 
  We calculated Euclidean distance between each point and normalized it by the maximum distance.
  The topological complexity of both image spaces was compared by the change of Betti numbers in Fig.~\ref{fig:topo_exp}, which 
 clearly showed that the manifold of the residual images are topologically simpler. 
 Indeed, $\beta_0$  of residual image manifold decreased faster to a single cluster.
 Moreover, there exists no $\beta_1$ barcode for the residual manifold which infers a closely distributed point clouds as spherical example in Fig.~\ref{fig:topo}(a)(b).
These results clearly informs  that the residual image manifold has a simpler topology 
  than the original one.

\begin{figure}[!hbt]
    \centerline{\includegraphics[width=0.8\linewidth]{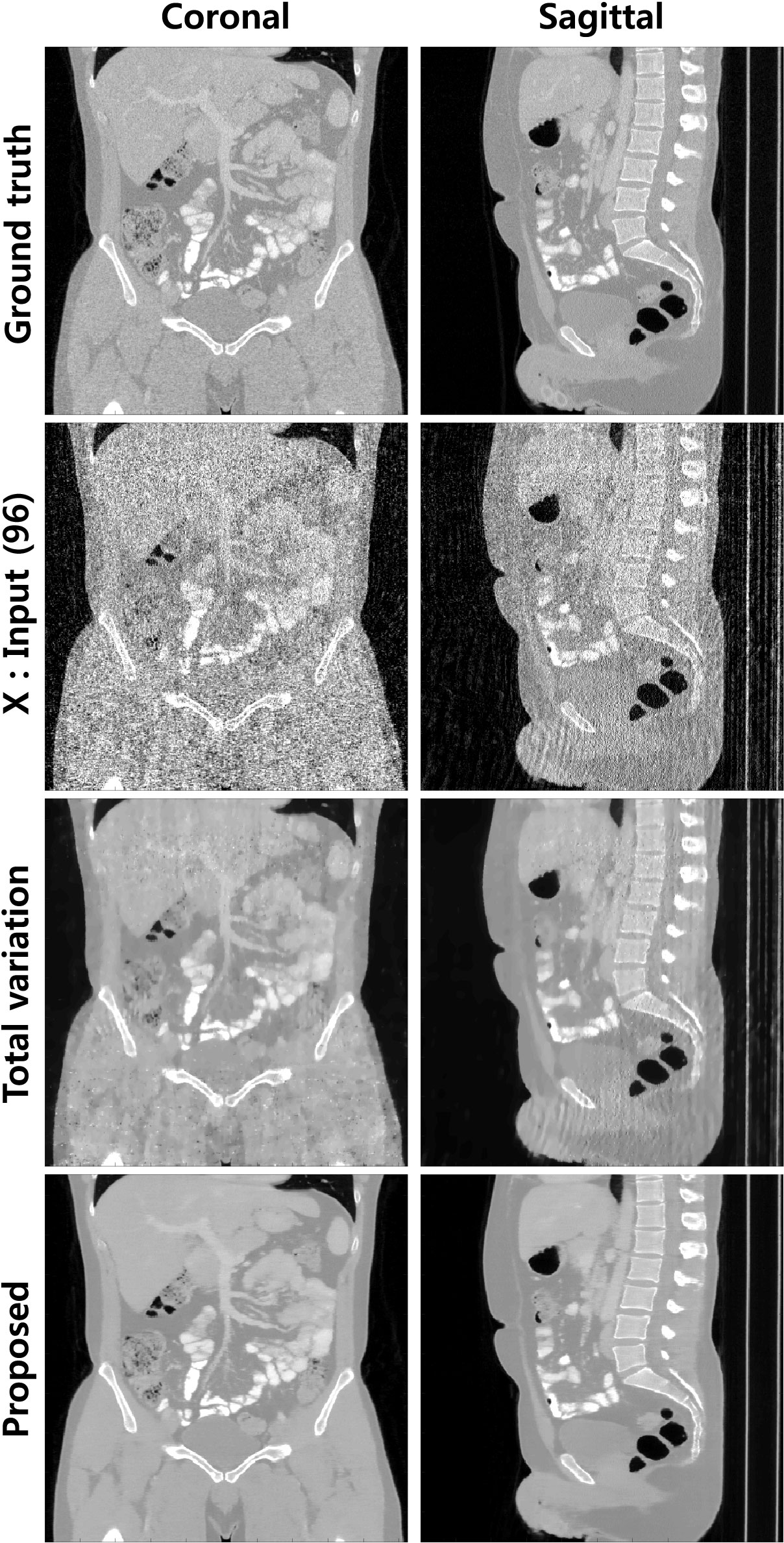}}
    \caption{Reconstruction results by TV based compressed sensing CT, and the proposed method.}
    \label{fig:cutview_result}
\end{figure}

\begin{figure*}[!hbt]
    \centerline{\includegraphics[width=0.95\linewidth]{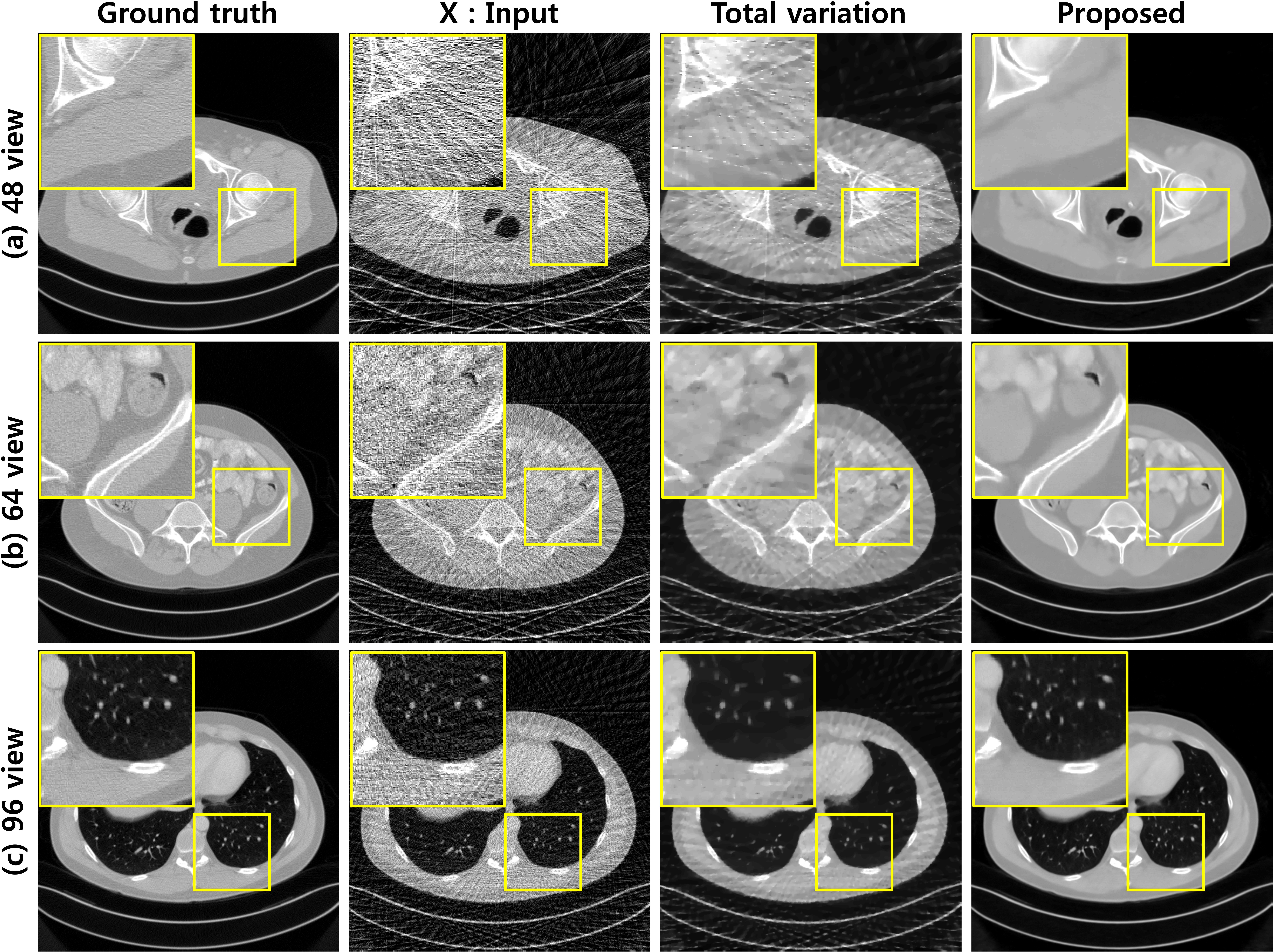}}
    \caption{Axial view reconstruction results by TV based compressed sensing CT, and the proposed method.}\label{fig:proposed_result}
\end{figure*}

\subsection{Network  training}
The proposed network was trained by stochastic gradient descent (SGD).  The regularization parameter was $\lambda = 10^{-4}$. The learning rate was set from $10^{-3}$ to $10^{-5}$ which was gradually reduced at each epoch. The number of epoch was 150. A mini-batch data using image patch was used,  and the size of image patch was $256\times256$. 
 
The network was implemented using MatConvNet toolbox (ver.20) \cite{vedaldi2015matconvnet} in MATLAB 2015a environment (MathWorks, Natick). We used a GTX 1080 graphic processor and i7-4770 CPU (3.40GHz). The network takes about 1 day for training.


\subsection{Reconstruction results}

Fig. \ref{fig:cutview_result} shows reconstruction results from coronal and sagittal directions. Accurate reconstruction was obtained using the proposed method, whereas there exist remaining patterned artifacts in TV reconstruction.
Fig. \ref{fig:proposed_result}(a)-(c) shows the reconstruction results from axial views the proposed methods from 
48, 64, and 96 projection views, respectively. 
Note that the same network was used for all these cases to verify the universality of the proposed method.
The results  in Fig. \ref{fig:proposed_result}(a)-(c) clearly showed that the proposed network removes most of streaking artifact patterns and  preserves a detailed structure of underlying images. 
The magnified view in Fig. \ref{fig:proposed_result}(a)-(c) confirmed that the detailed structures are very well reconstructed using the proposed method.
Moreover, compared to the standard compressed sensing CT approach with TV penalty, the proposed results in Fig.~\ref{fig:cutview_result} and  Fig.~\ref{fig:proposed_result}
provides significantly improved image reconstruction results, even though the computational time for the proposed method is 123ms/slice.
This is 30 time faster  than the TV approach where the standard TV approach took
about 3 $\sim$ 4 sec/slice for reconstruction.


\section{Discussion}


\subsection{Residual learning vs. Image learning}

\begin{figure}[!hbt]
    \centerline{\includegraphics[width=0.95\linewidth]{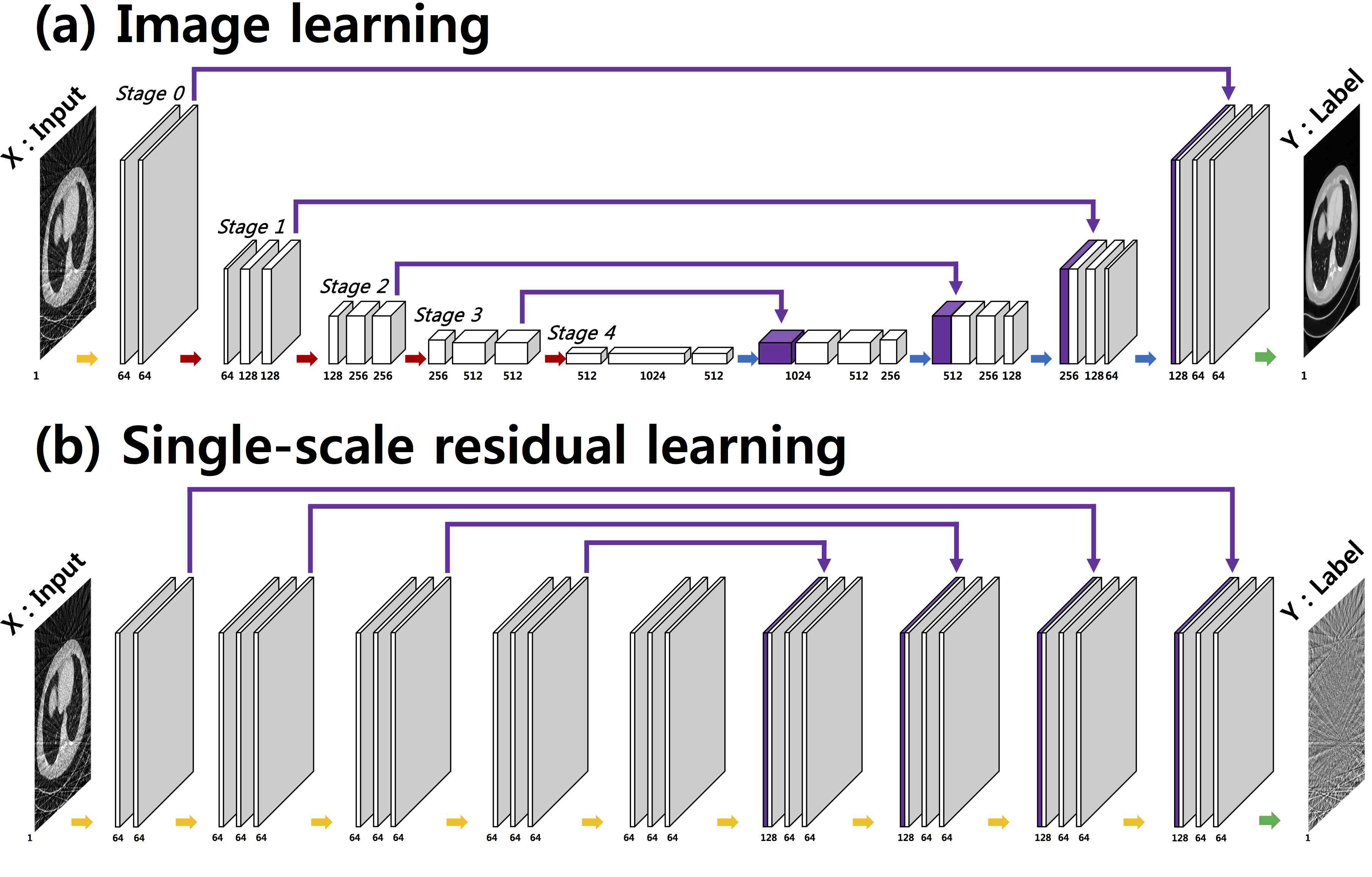}}
    \caption{Reference networks.}
    \label{fig:compared_network}
\end{figure}

Here, we conduct various comparative studies.
First,  we investigated the importance of the residual learning. 
As for reference, an image learning network in Fig.~\ref{fig:compared_network}(a) was used.
Although this has the same U-net structure as the proposed residual learning network in Fig.~\ref{fig:proposed_network}, the full view reconstruction results
were used as label and the network was trained to learn the artifact-free images.
According to  our persistent homology analysis, the manifold of full-view reconstruction is topologically more complex so that the learning the full
view reconstruction images is more difficult.

\begin{figure}[!hbt]
	\centering
    \begin{minipage}[b]{0.45\linewidth}
			\centerline{\includegraphics[width=\linewidth]{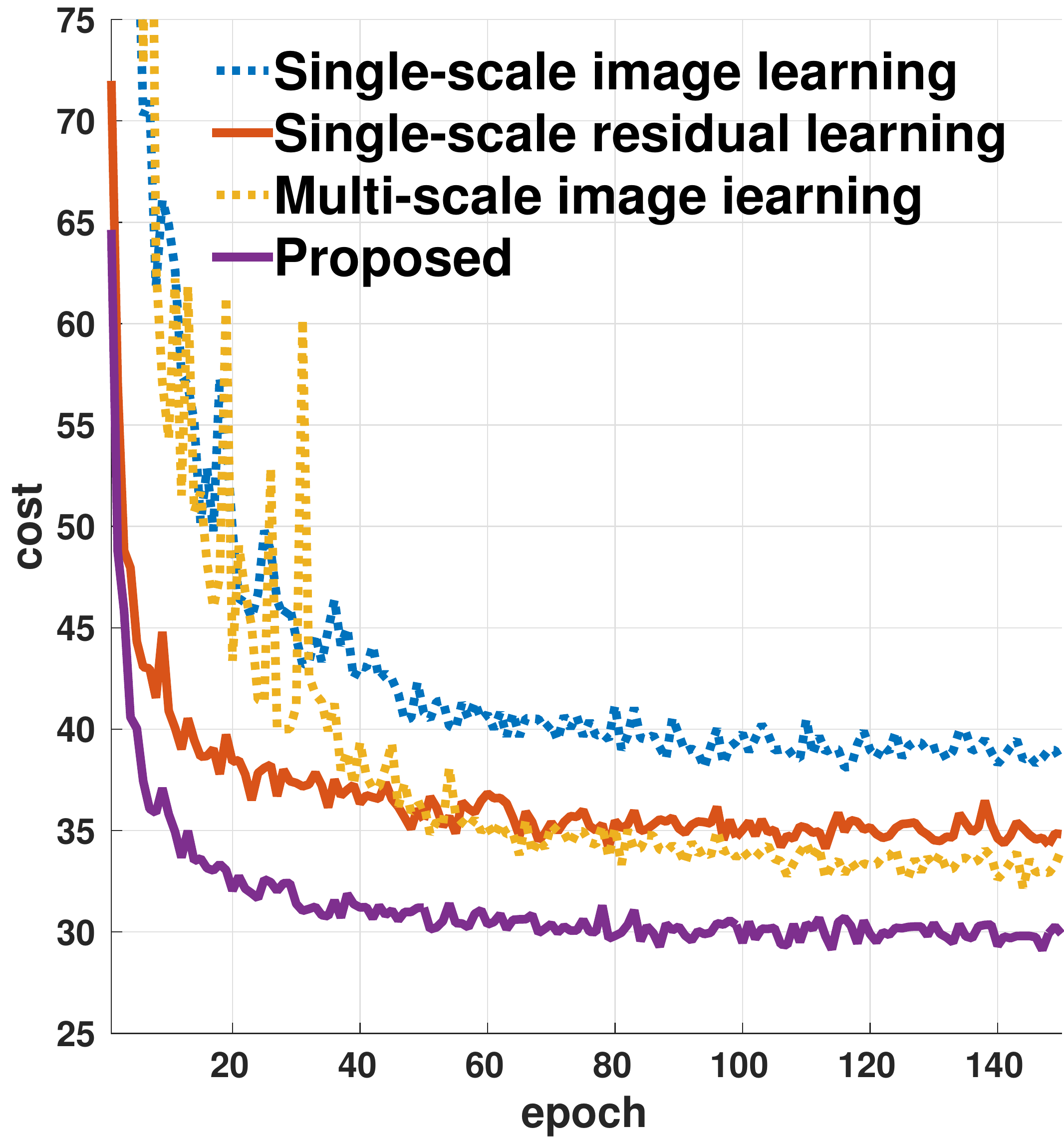}}
			\centerline{(a) COST}\medskip
    \end{minipage}
    \begin{minipage}[b]{0.45\linewidth}
			\centerline{\includegraphics[width=\linewidth]{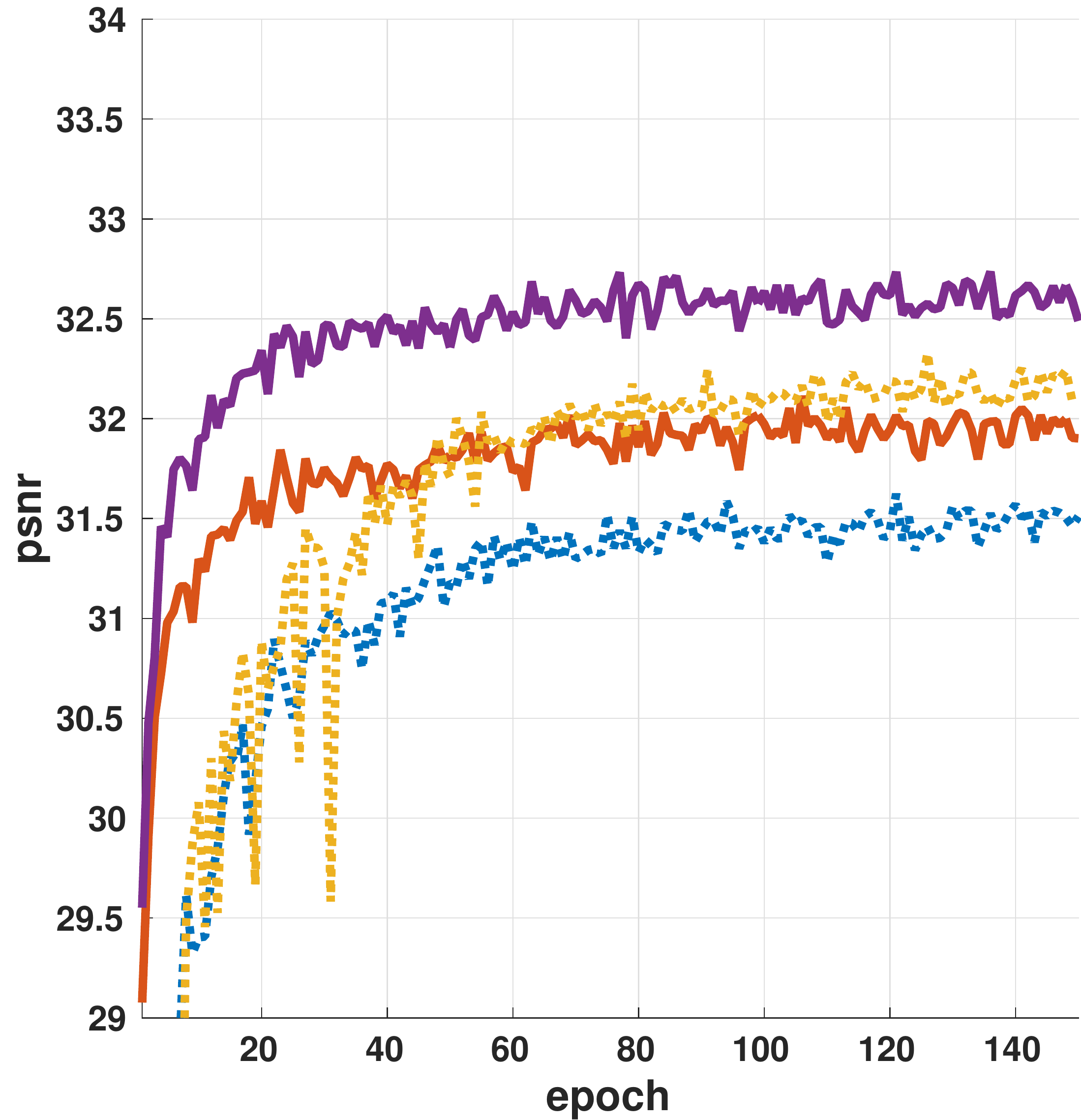}}
			\centerline{(b) PSNR}\medskip
    \end{minipage}
	\caption{Convergence plots for  (a) cost function, and  (b) peak-signal-to-noise ratio (PSNR) with respect to
	 each epoch.}
	\label{fig:err_plot}
\end{figure}

\begin{figure*}[!hbt]
    \centerline{\includegraphics[width=0.95\linewidth]{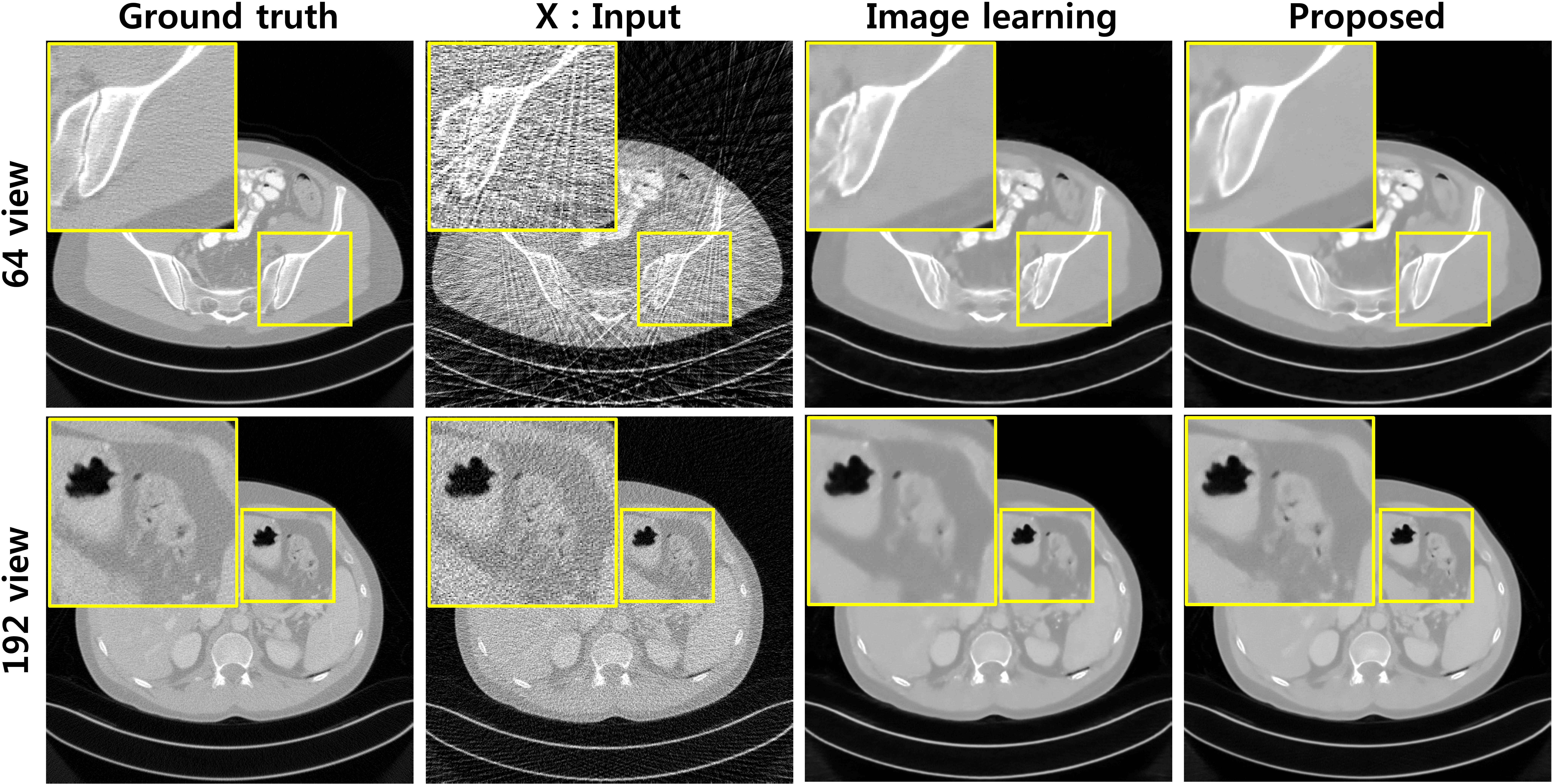}}
    \caption{Comparison results for residual learning and image learning from 64 and 192 view reconstruction input data. }
    \label{fig:image_result}
\end{figure*}

The convergence plot in Fig.~\ref{fig:err_plot} and reconstruction results in Fig. \ref{fig:image_result}  clearly show the strength of the residual learning over the image learning.
The proposed residual learning network exhibits the fast convergence during the training
and the final performance  outperforms the image learning network.
The magnified view  of a reconstructed image in Fig. \ref{fig:image_result} clearly shows that
the detailed structure of internal organ was not fully recovered using image learning, whereas the proposed residual learning can recover.

\subsection{Single-scale vs. Multi-scale residual learning}

Next,  we investigated the importance of the multi-scale nature of residual learning using U-net. 
As for reference, a single-scale residual learning network as shown in Fig.~\ref{fig:compared_network}(b) was used.
Similar to the proposed method,
the streaking artifact  images were used as the labels. However, the residual network was constructed without pooling and unpooling layers.
For fair comparison, we set the number of network parameters similar to the proposed method by fixing the number of channels at each layer across all the stages.  

\begin{figure*}[!hbt]
    \centerline{\includegraphics[width=0.95\linewidth]{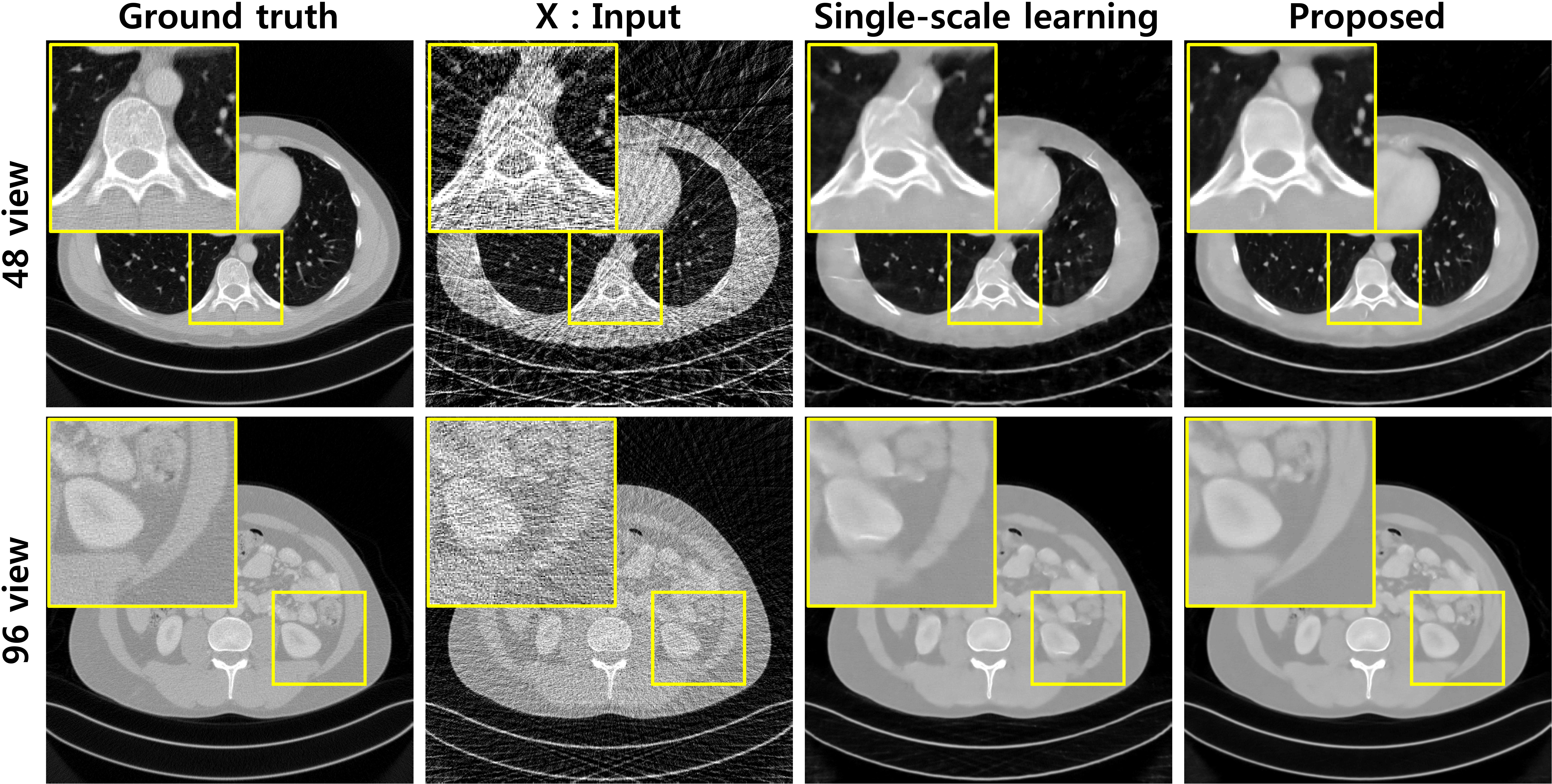}}
    \caption{Comparison results for single-scale versus multi-scale residual  learning  from 48 and 96 view reconstruction input data. }
    \label{fig:single_scale_result}
\end{figure*}

Fig. \ref{fig:err_plot} clearly shows the advantages of multi-scale residual learning over 
single-scale residual learning.
The proposed residual learning network exhibits the fast convergence and the final performance was better.
In Fig.~\ref{fig:single_scale_result},  the image reconstruction quality by the multi-scale learning was much improved compared to the 
single resolution one.

PSNR is shown as the quantification factor Table. \ref{tlb:err_table}.
Here, in extremely sparse projection views, multi-scale structures always win 
 the single-scale residual learning.
At 192 projection views, the global streaking artifacts become less dominant compared to the localized artifacts,
so the single-scale residual learning started to become better than multi-scale {\em image} learning approaches.
However,  by combining the advantages of residual learning and multi-scale network,
 the proposed multi-scale residual learning outperforms all the reference architectures in various view downsampling ranges.

\subsection{Diversity of training set}

\begin{figure*}[!hbt]
    \centerline{\includegraphics[width=0.95\linewidth]{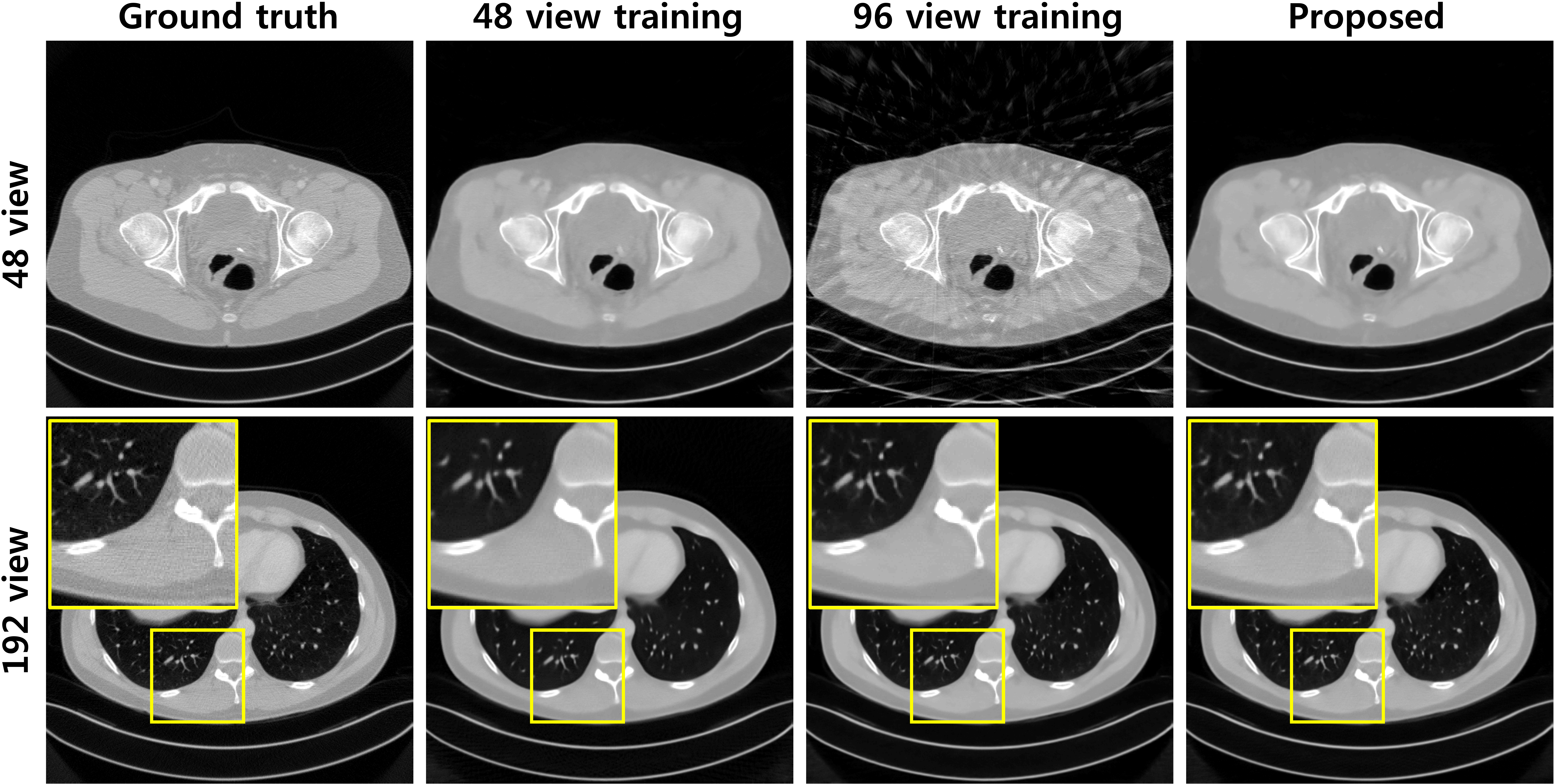}}
    \caption{Comparison results for various training data configuration. Each column represents reconstructed images by proposed network which was trained with sparse view reconstruction from 48, 96, or 48/96 views, respectively.}
    \label{fig:training_case_result}
\end{figure*}

Fig. \ref{fig:training_case_result} shows that reconstructed results by the proposed approach,
when the network was trained with sparse view reconstruction from 48, 96, or  48 and 96 views, respectively. 
The streaking artifacts were removed very well in all cases; however, the detailed structure of underlying image was maintained  when the network was trained
using 96 view reconstruction, especially when  the network was used to reconstruct image from more dense view data set (198 views in this case).
%
On the other hand,  the network trained with 96 views could not be used for 48 views sparse CT reconstruction.
Fig. \ref{fig:training_case_result} clearly showed the remaining streaking artifact in this case.

To address this issue and make the network universal across wide ranges of view down-sampling,
 the proposed network was, therefore,  trained using a training data set by combining sparse CT reconstruction data between 48 and 96 view. 
 As shown in Fig. \ref{fig:training_case_result},  the proposed approach with the combined training provided the best reconstruction across wide
 ranges of view down-sampling.

\begin{table*}
	\begin{center}
		\begin{tabular}{|c|c|c|c|c|c|}
			\hline
			 \  No. of views & Single-scale image learning & Single-scale residual learning & Multi-scale image learning & Proposed \\
			\hline\hline
			48 view		& 31.0027 								& 31.7550	& $\color{blue}{32.5525}$	& $\color{red}{\bold{33.3916}}$ \\
			64 view		& 32.1380 								& 32.4456	& $\color{blue}{32.9748}$ & $\color{red}{\bold{33.8680}}$ \\
			96 view		& 33.2983 								& 33.3569	& $\color{blue}{33.4728}$ & $\color{red}{\bold{34.5898}}$ \\
			192 view	& $33.7693$								& \color{blue}{33.8390}	& 33.7101		& $\color{red}{\bold{34.9028}}$ \\
			
			
			
			\hline
		\end{tabular}
	\end{center}
	\caption{Average PSNR results for various learning architectures.}
	\label{tlb:err_table}
\end{table*}

\section{Conclusion}

In this paper, we develop a novel deep residual learning approach for sparse view CT reconstruction.
Based on the persistent homology analysis, we showed that the residual manifold composed of streaking artifacts is topologically simpler than the original one.
This claim was confirmed using persistent homology analysis
and experimental results, which clearly showed the advantages of the residual learning over image learning.
Among the various residual learning networks, this paper showed that the multi-scale residual learning using U-net structure was the most effective especially when the
number of views were extremely small. 
We showed that this is due to the enlarged receptive field in U-net structure that can easily capture the globally distributed
streaking artifacts.
Using extensive experiments, we showed that the proposed deep residual learning is significantly better than the conventional compressed sensing CT approaches.
Moreover, the computational speed was extremely faster than that of compressed sensing CT.

Although this work was mainly developed for sparse view CT reconstruction problems,  the  proposed residual learning network may be universally used for removing various image
noise and artifacts that are globally distributed. 

\section{Acknowledgement}

The authors would like to thanks Dr. Cynthia MaCollough,  the Mayo Clinic, the American Association of Physicists in Medicine (AAPM), and grant EB01705 and EB01785 from the National
Institute of Biomedical Imaging and Bioengineering for providing the Low-Dose CT Grand Challenge data set.
This work is supported by Korea Science and Engineering Foundation, Grant number
NRF-2016R1A2B3008104.

%


%

\end{document}